# Double-Shot 3D Shape Measurement with a Dual-Branch Network for Structured Light Projection Profilometry

Mingyang Lei, Jingfan Fan*, Long Shao*, Hong Song, Deqiang Xiao, Danni Ai, Tianyu Fu, Yucong Lin, Ying Gu, and Jian Yang*

***Abstract*—The structured light (SL)-based three-dimensional (3D) measurement techniques with deep learning have been widely studied to improve measurement efficiency, among which fringe projection profilometry (FPP) and speckle projection profilometry (SPP) are two popular methods. However, they generally use a single projection pattern for reconstruction, resulting in fringe order ambiguity or poor reconstruction accuracy. To alleviate these problems, we propose a parallel dual-branch Convolutional Neural Network (CNN)-Transformer network (PDCNet), to take advantage of convolutional operations and self-attention mechanisms for processing different SL modalities. Within PDCNet, a Transformer branch is used to capture global perception in the fringe images, while a CNN branch is designed to collect local details in the speckle images. To fully integrate complementary features, we design a double-stream attention aggregation module (DAAM) that consists of a parallel attention subnetwork for aggregating multi-scale spatial structure information. This module can dynamically retain local and global representations to the maximum extent. Moreover, an adaptive mixture density head with bimodal Gaussian distribution is proposed for learning a representation that is precise near discontinuities. Compared to the standard disparity regression strategy, this adaptive mixture head can effectively improve performance at object boundaries. Extensive experiments demonstrate that our method can reduce fringe order ambiguity while producing high-accuracy results on self-made datasets.**

## I. INTRODUCTION

THE development of information technology has accelerated human life into the three-dimensional (3D) world [1-5]. Among various measurement technologies,

This work was supported in part by National Science Foundation Program of China (Grant 62025104, 62302042, 62422102), National Key Reasearch and Development Program of China (Grant 2023YFC2415300), China Postdoctoral Science Foundation (Grant 2024M754100). *(Corresponding author: Jingfan Fan, Long Shao, and Jian Yang).*

Mingyang Lei, Tianyu Fu and Ying Gu are with the School of Medical Technology, Beijing Institute of Technology, Beijing 100081, China (e-mail: 13120030055@163.com; fty0718@bit.edu.cn; guyinglaser301@163.com).

Long Shao and Hong Song is with the School of Computer Science and Technology, Beijing Institute of Technology, Beijing 100081, China (e-mail: songhong@bit.edu.cn).

Jingfan Fan, Deqiang Xiao, Danni Ai, Yucong Lin, and Jian Yang are with the School of Optics and Photonics, Beijing Institute of Technology, Beijing 100081, China (e-mail: fjf@bit.edu.cn; xiaodq2011@gmail.com; danni@bit.edu.cn; linyucong@bit.edu.cn; jyang@bit.edu.cn).

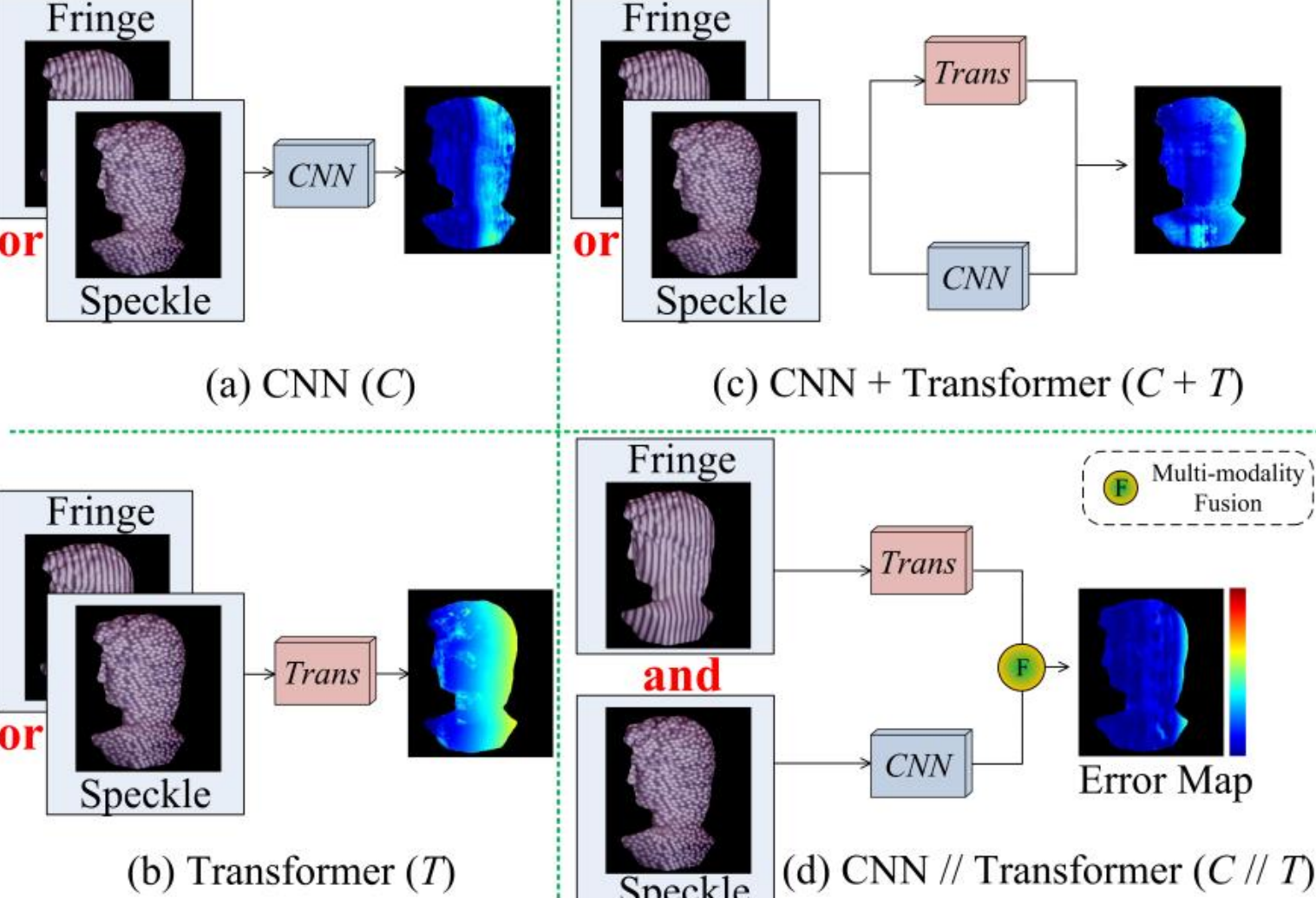

Fig. 1. Existing SL-based measurement methods using deep learning ((a) FOA-Net [9], (b) CTransU-Net [10], and (c) PCTNet [11]) vs. Our PDCNet. Note that we show the results of the best-performing patterns for these comparative methods.

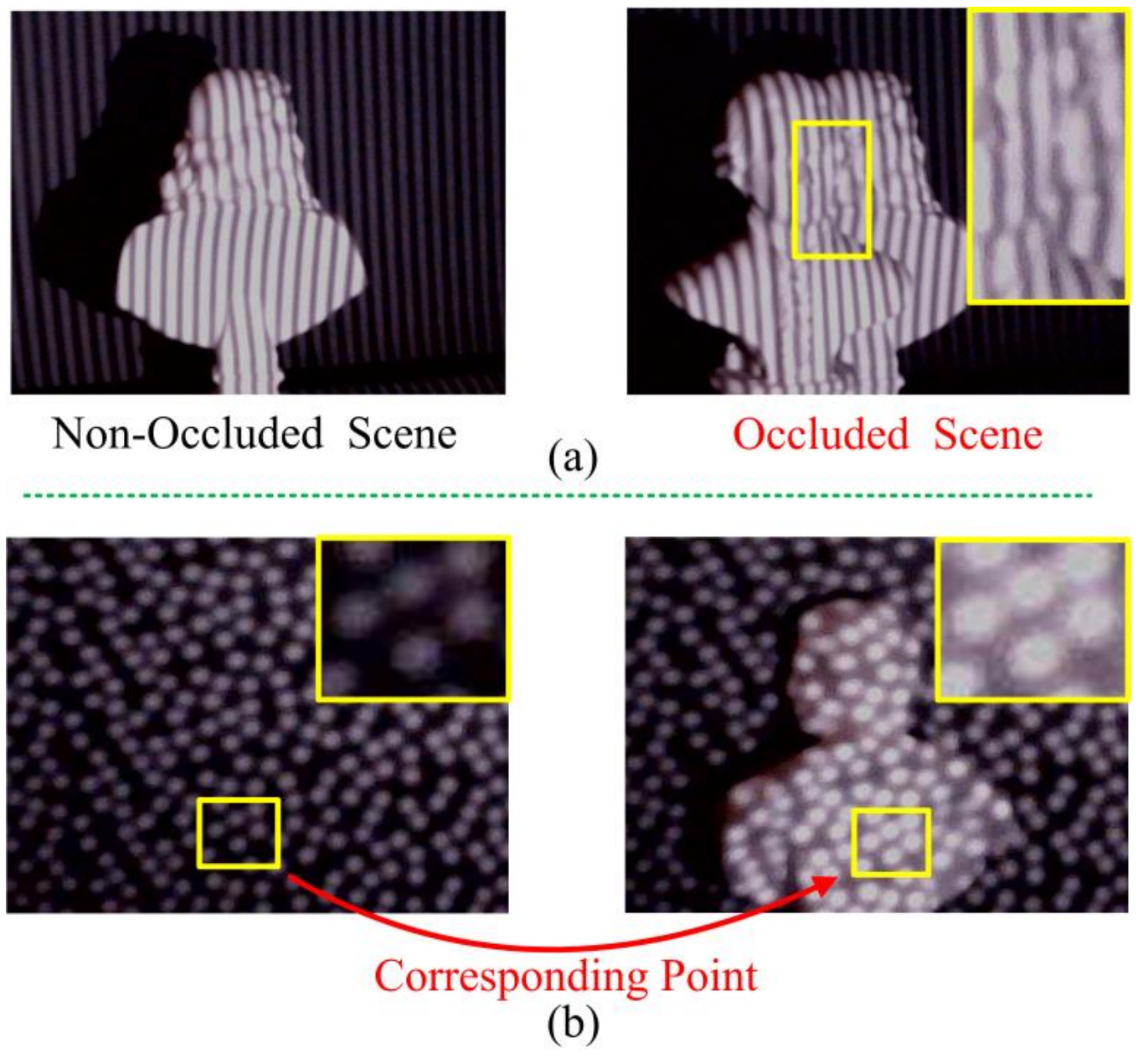

Fig. 2. The characteristics of images with different SL patterns. (a): Fringe images. (b): Speckle images.

structured light (SL) projection profilometry has become a popular non-contact 3D shape measurement technique. Over the past few decades, the traditional methods using fringe patterns and speckle patterns have been regarded as the standard practice for various applications [6, 7]. However, these traditional methods generally require multiple input images and manually designed algorithms (such as phase unwrapping and speckle matching) to obtain depth information. To improve measurement efficiency, many researchers have leveraged the powerful representation capabilities of deep learning networks to perform direct end-to-end measurements by inputting a small number of SL images [8]. Among these methods, fringe projection profilometry (FPP) based on sinusoidal fringe patterns, and speckle projection profilometry (SPP) based on speckle patterns are two popular methods. However, as shown in Fig. 1, they are all single-shot-based methods [9-11], which use a fringe or a speckle image as input and may cause errors in actual measurements. To be specific, FPP methods generally obtain the 3D measurement results from a few fringe patterns. Due to the continuity and periodicity nature of sinusoidal patterns, determining accurate fringe order with global information is the key to pixel-wise phase measurement. However, in some occluded scenes, the change in depth may lead to ambiguity in the fringe order. As illustrated by the yellow box in Fig. 2(a), the order of the fringes projected on two objects is difficult to judge. As for the SPP methods, the speckle images modulated by the object surface are captured by the camera and then utilized to obtain the depth maps with various networks. However, SPP technology does not employ pixel-by-pixel encoding [12]. Instead, it mainly encodes the feature information within the local speckle regions, lacking pixel-wise global information like that within the fringe image, which may obtain poor results. As shown in Fig. 2(b), for a monocular SL system, the change in depth will bring about the movement of the speckle spots, but the distribution pattern of these spots remains unchanged. Therefore, the local information within the speckle images is the most significant clue for measurement.

To fully leverage the advantages of these deep learning-based methods and alleviate the aforementioned problems in SL-based 3D measurement tasks, we propose a dual-branch Convolutional Neural Network (CNN)-Transformer framework for recovering accurate 3D shapes (Fig. 1(d)). Within our double-shot-based method, we use the convolution operators to extract local features from speckle images and the self-attention mechanisms to capture global representations from fringe images. The main contributions of our paper are as follows:

(1) Considering the complementarity of the fringe and the speckle images, we feed them into separate Transformer and CNN branches to capture global perception and local details respectively. Especially in the fringe ambiguity regions, the distribution invariance of speckles helps PDCNet to achieve excellent measurement performance.

(2) We propose a double-stream attention aggregation module (DAAM) to integrate complementary features, which can aggregate multi-scale structure information. Especially in the occluded scenes, DAAM pays more attention to crucial information compared to other attention modules.

(3) For recovering sharp boundaries, we exploit a bimodal Gaussian distribution with disparity adaptive selection strategy, to learn a representation that is precise near discontinuities.

We validate the effectiveness of the proposed method on a self-made dataset, which consists of occluded scenes and non-occluded scenes. Experiments show that compared with other deep learning models, our method can generate higher precision results while reducing fringe ambiguity.

## II. Related Works

### *A. Traditional SL-based measurement*

In recent decades, the traditional SL-based measurement methods utilizing fringe patterns and random speckle patterns have been adopted as the standard practice for many applications [13, 14]. In the traditional FPP-based methods [15-18], the sinusoidal phase-shifting algorithm is a popular branch in them because of its high resolution and robustness. In this algorithm, $N$ standard sinusoidal fringe patterns are projected on the object, whose intensities can be expressed as follows:

$$I_i^p = a + b\cos(2\pi\frac{x}{T} + 2\pi\frac{i}{N}),\quad i = 0,1,...,N-1 \tag{1}$$

where the superscript letter $p$ refers to the projector. $a$ is the background intensity of the pattern, and $b$ is the modulation intensity of the fringe. $T$ denotes the period length. The captured deformed fringe images by the camera are expressed by the following equation:

$$I_i^c = A + B\cos(\varphi + 2\pi\frac{i}{N}),\quad i = 0,1,...,N-1 \tag{2}$$

where the superscript letter $c$ refers to the camera. $A$ and $B$ represent the background intensity and the modulation intensity respectively. $\varphi$ is the absolute phase information, which is retrieved by the following:

$$\psi = \arctan\frac{\sum_{i=0}^{N-1} I_i^c \sin(2\pi i/N)}{\sum_{i=0}^{N-1} I_i^c \cos(2\pi i/N)},\quad i = 0,1,..., N-1 \tag{3}$$

Note that the wrapped phase $\psi$ is wrapped in the range $[-\pi, +\pi]$ with $2\pi$ discontinuities. Therefore, a phase unwrapping method should be used to generate the absolute phase. Fig. 3 shows the generation process of the wrapped phase and the absolute phase.

In the traditional SPP-based methods [7, 19, 20], the speckle image of the object $I_{si}$ is matched with the reference speckle image $I_{re}$, which is captured when the camera's optical axis is perpendicular to a planar object at a distance $Z_{ref}$. Then, the corresponding relationships between them are computed with a block matching-based method $BM(\cdot)$ to generate the disparity map $d_m$

$$d_m = BM(I_{si}, I_{re}) \tag{4}$$

Finally, the depth map $Z_m$ can be generated via

$$Z_m = \frac{Z_{ref}}{1 - \frac{Z_{ref} d_m}{B_m f_m}} \tag{5}$$

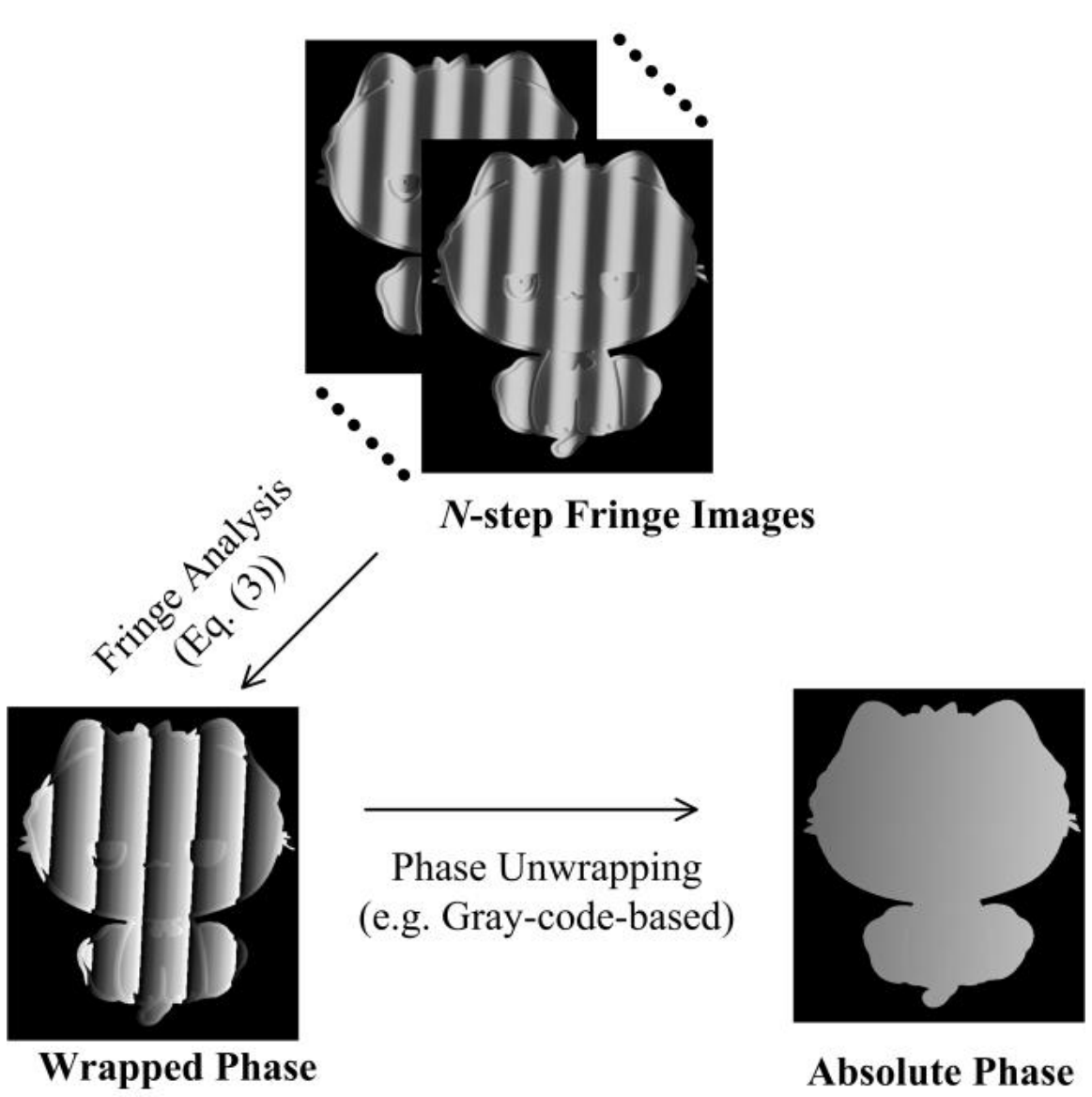

Fig. 3. The generation process of the wrapped phase and the absolute phase.

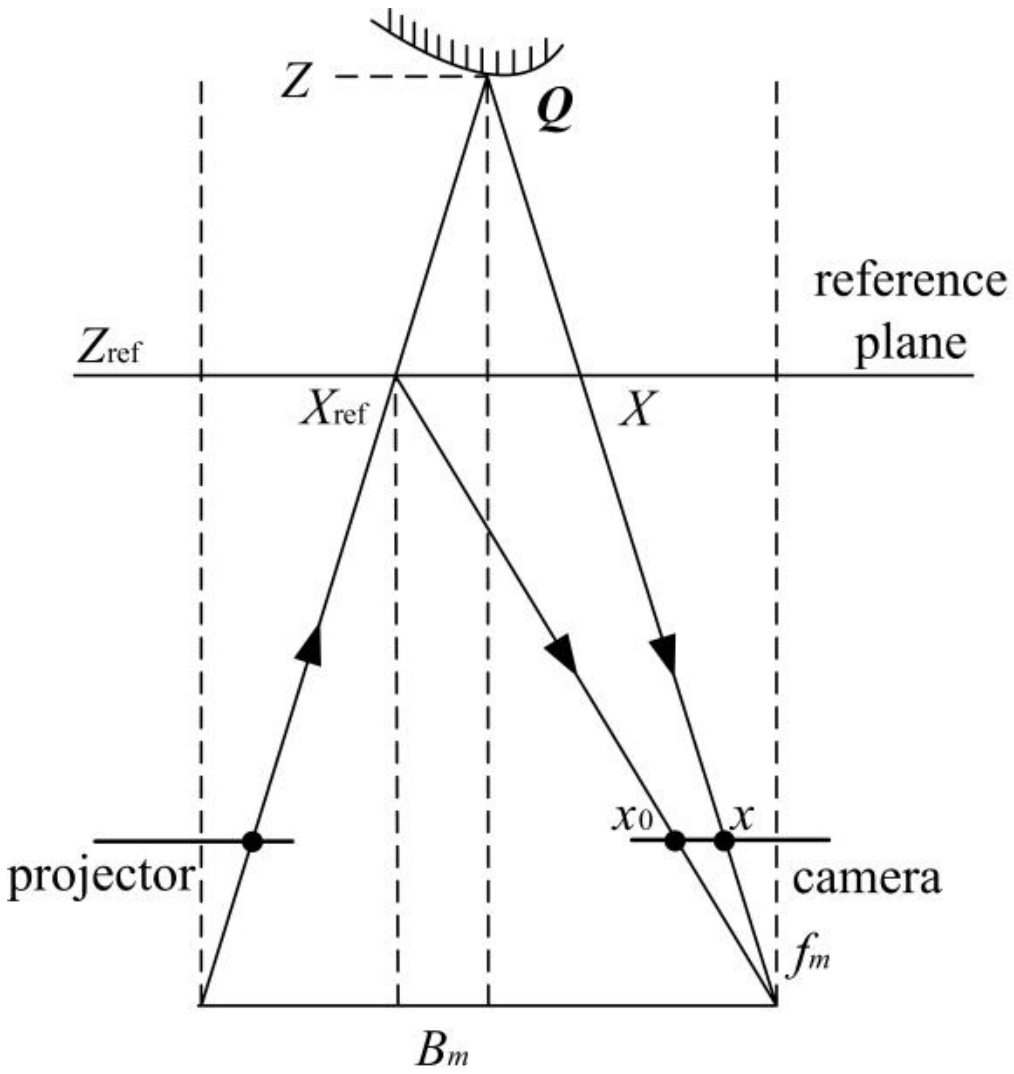

Fig. 4. Geometry of the SL system.

where $f_m$ and $B_m$ denote the focal length and the baseline of the SL system. Fig. 4 shows the geometry of the SL system.

*B. SPP-Based Measurement using deep learning*

To improve measurement efficiency, analyzing speckle patterns using deep learning has become popular in recent years. By combining photometric and geometric information, Riegler et al. [21] posed depth estimation as a regression task, which trains a neural network in a self-supervised manner. Leveraging optical flow, DepthInSpace [22] used information from multiple video frames from the same scene to boost depth estimation accuracy in three different self-supervised ways. Nguyen et al. [23] transformed a single shot image with a dense random speckle pattern, into a 3D shape using a deep CNN framework, which has a contracting encoder path and an expansive decoder path. In addition, for a monocular structured light system, establishing the correspondences between the current and the reference speckle images is the key to recovering 3D shapes. Therefore, many binocular speckle stereo-matching frameworks can also be adopted for this task. Zhang et al. [24] presented ActiveStereoNet to cope with illumination effects and occluded pixels, which is the first deep-learning method for active stereo systems. Liu et al. [25] presented a lightweight stereo-matching algorithm through unsupervised learning, which realizes real-time stereo-matching with low power consumption. However, SPP technology relies on local spatial features but neglects pixel-level information, resulting in low measurement accuracy.

*C. FPP-Based Measurement using deep learning*

Due to the need for high-precision reconstruction, many recent deep learning-based approaches utilize FPP for measurement. Wang et al. [26] proposed a lightweight dual-path hybrid network with a multi-attention mechanism based on UNet [27], which reduces the learnable parameters by 60% compared with the original framework. Guo et al. [9] proposed a generic temporal phase unwrapping framework, which can effectively improve the performance of phase unwrapping and is less sensitive to noise than traditional methods [28]. To exploit both spatial and temporal phase information in an integrated way, DL-TPU [29] achieved high-speed 3D surface imaging with the use of 6 projected patterns without exploring geometric constraints. Yu et al. [30] proposed an untrained deep learning-based phase retrieval method, which first transforms the captured fringes and coarse wrapped phases into desired fringe orders, and then optimizes these absolute phases with scene-independent physical constraints. However, these methods can not guarantee stable fringe order ambiguity removal in practical applications. To solve this problem, Li et al. [31] predicted the numerator term and the denominator term of the arctangent function of the wrapped phase. However, this method does not adopt an end-to-end network structure, which requires additional computation to obtain the absolute phase map.

*D. Combination of CNN and Transformer*

Recent works show the advantages of combining CNN and Transformer in different computer vision tasks [32-35]. Peng et al. [32] proposed a hybrid network Conformer, to take advantage of self-attention mechanisms and convolutional operations for boosting representation learning. Mobile-Former [33] presented a parallel design of MobileNet [36] and transformer to obtain more representation information. BoTNet [34] improved both instance segmentation and object detection by using self-attention in the last three blocks of ResNet [37]. ConViT [35] presented a gated positional self-attention (GPSA) to improve trainability and sample efficiency. Experimental results have consistently indicated that the hybrid models combining CNN and Transformer frameworks exhibit superior performance compared to using them individually.

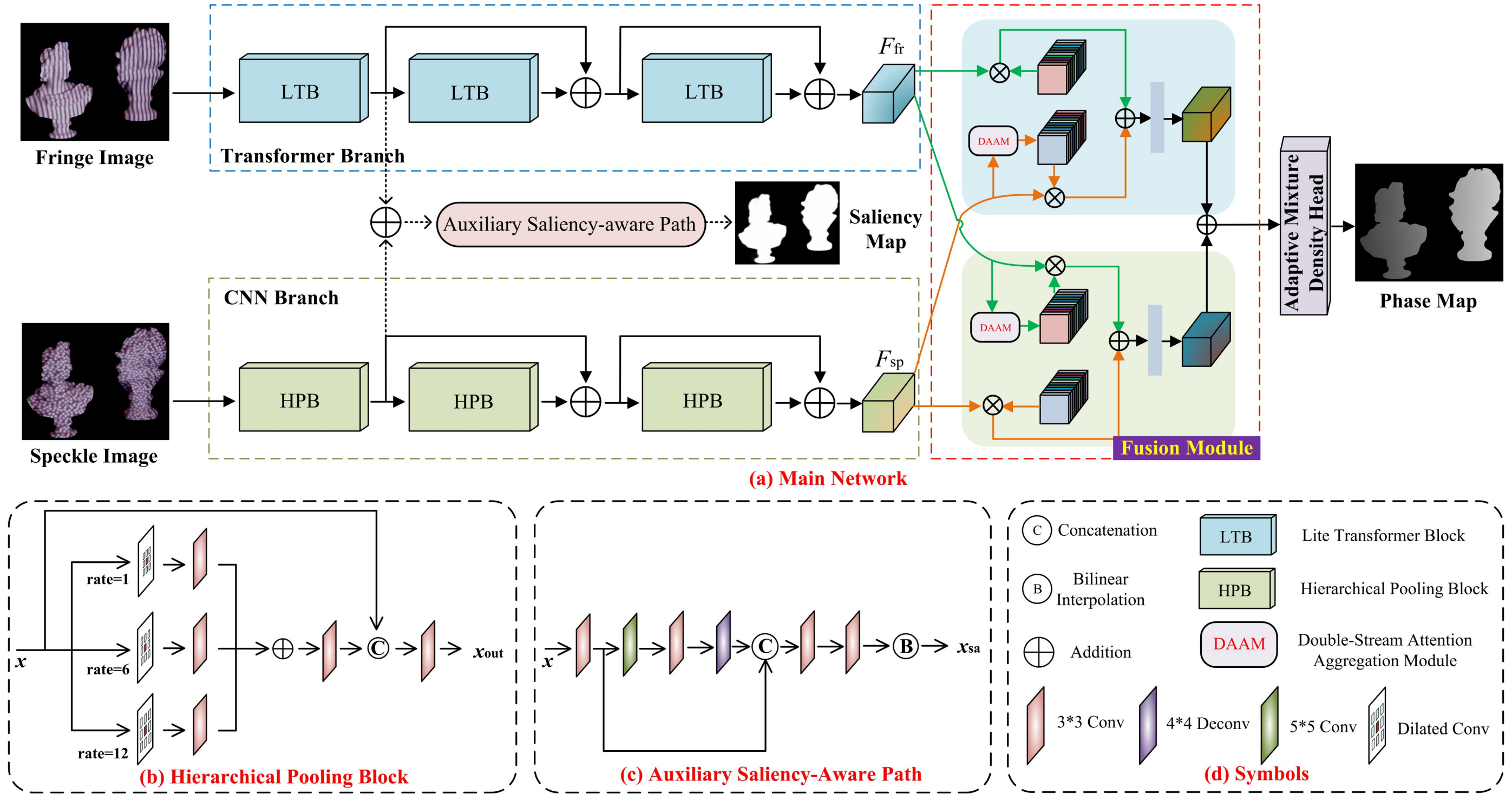

Fig. 5. The architecture of our PDCNet. PDCNet receives two images of different modalities as inputs and predicts the corresponding absolute phase map. The fusion module is used to integrate complementary features. The saliency-aware path is used to improve the semantic characteristic of the output phase map.

To tackle the problems of order ambiguity in fringe images and the absence of global information in speckle images, we propose an end-to-end dual-branch CNN-Transformer framework for extracting and fusing global and local features, which are extracted from fringe and speckle images respectively.

## III. Method

### A. Network Structure

In CNN, the convolution operations are good at extracting local features but difficult to capture global representations. In contrast, in Transformer, cascading attention mechanisms can model long-range dependencies, but break local feature details. To solve these problems, we propose a novel framework to take advantage of convolutional operations and self-attention modules for enhancing the representation learning of speckle images and fringe images respectively (Fig. 5).

### B. CNN Branch

The CNN branch consists of three hierarchical pooling blocks (HPB) for effective incorporation of context. The HPB (Fig. 5(b)) first uses dilation convolution with different dilation rates and a 3×3 convolution layer to form a pyramid feature, which can enlarge the receptive fields with high efficiency. Then, we add up the obtained features of different branches, and a convolution layer is followed to get the aggregated feature map. Finally, a dense connection is used for generating the final output feature map $x_{out}$, which has a regularizing effect and can reduce overfitting on the phase measurement task. The final output feature map can be summarized as:

$$x_{out} = f^{3\times3}([f^{3\times3}(f^{3\times3}(r_1(x)) \oplus f^{3\times3}(r_6(x)) \oplus f^{3\times3}(r_{12}(x))); x]) \tag{6}$$

where $r_i$ represents the convolution kernel with dilation factor $i$. $f^{3\times3}$ denotes a convolution operation with a filter size of 3×3.

### C. Transformer Branch

The Transformer branch contains two stages: Pre-feature extraction and Main feature extraction. The previous work by Tete et al. [38] demonstrated that convolution is suitable for early visual processing. Therefore, we use a 3 × 3 convolution layer ($H_{pre}(\cdot)$) for this stage:

$$F_{pre} = H_{pre}(I_{fr}) \tag{7}$$

Then, the main feature is extracted from $F_{pre}$:

$$F_{main} = H_{main}(F_{pre}) \tag{8}$$

Considering to balance the performance and computational efficiency, we use three Lite Transformer blocks (LTB) [39] as the basic unit ($H_{main}(\cdot)$), which shrinks the embedding size to reduce the total computation amount while maintaining the same performance.

### D. Auxiliary saliency-aware path

It has been verified by previous studies [40] that a suitable auxiliary constraint path is beneficial to improving the performance of different related tasks. Furthermore, similar tasks can reinforce each other, such as semantic labels and depth maps [41], which are both pixel-level and inherently interconnected, can mutually reinforce each other. Therefore, to boost the semantic characteristic of the output absolute phase map, the saliency map and the absolute phase map are simultaneously predicted in our proposed PDCNet.

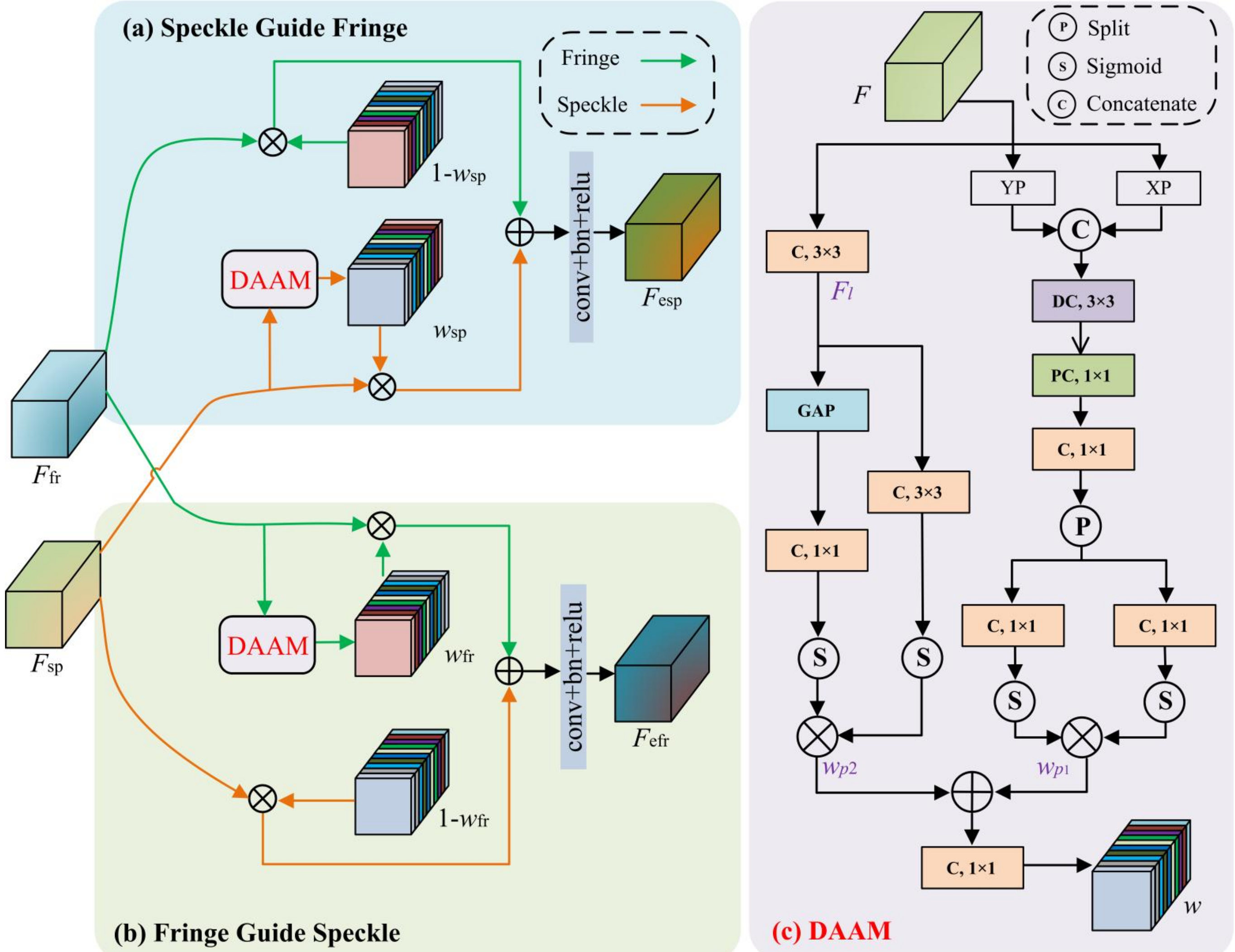

Fig. 6. The proposed fusion strategy and detailed structure of DAAM. 'XP' and 'YP' denote 1D horizontal global pooling and 1D vertical global pooling, respectively. 'DC': Depthwise convolution. 'PC': Pointwise convolution. 'GAP': Global average pooling. 'C': Convolution. ⊕ : Element-wise sum. ⊗ : Elment-wise multiplication.

In our algorithm, we use an auxiliary saliency-aware path for boosting the performance of 3D shape measurement, which allows our PDCNet to focus more on object regions and less on background regions. As shown in Fig. 5(c), the main process can be described as:

$$x_{\mathrm{fi}} = f^{3\times3}(x) \tag{9}$$

$$x_{\mathrm{sa}} = Bi(f^{3\times3}(f^{3\times3}([d^{4\times4}(f^{3\times3}(f^{5\times5}(x_{\mathrm{fi}})));x_{\mathrm{fi}}]))) \tag{10}$$

where $x_{\mathrm{sa}}$ is the output saliency map. $f^{3\times3}$ and $f^{5\times5}$ represent the $3\times3$ and $5\times5$ convolution layer respectively. $d$ represents the deconvolution layer. $Bi(\cdot)$ denotes the bilinear interpolation.

In this paper, we use GCPANet [42] to generate the ground-truth saliency maps. Note that the auxiliary path only exists during training, whereas it is abandoned when inference.

*E. Fusion Strategy*

Fig. 6(a) and Fig. 6(b) show the specific structure of our cross-modal fusion strategy. Two weight matrices generated from $F_{\mathrm{sp}}$ and $F_{\mathrm{fr}}$ are multiplied with the input, to pay more attention to the important information and suppress the redundant features:

$$F_{\mathrm{esp}} = F_{\mathrm{sp}} * w_{\mathrm{sp}} + F_{\mathrm{fr}} * (1 - w_{\mathrm{sp}}) \tag{11}$$

$$F_{\mathrm{efr}} = F_{\mathrm{fr}} * w_{\mathrm{fr}} + F_{\mathrm{sp}} * (1 - w_{\mathrm{fr}}) \tag{12}$$

Then, these results are further added to generate the cross-modal feature $F_{\mathrm{cm}}$:

$$F_{\mathrm{cm}} = F_{\mathrm{esp}} + F_{\mathrm{efr}} \tag{13}$$

In this paper, we propose a novel double-stream attention aggregation module (DAAM) to obtain the weight matrix. Figure 6(c) presents the architecture of DAAM. Motivated by coordinate attention [43], we propose a parallel attention subnetwork for aggregating multi-scale spatial structure information.

Given a feature map $F$, a one-dimensional feature-encoding operation is performed on each channel using pooling kernels of size ($H$, 1) and (1, $W$) along the horizontal coordinate and the vertical coordinate, respectively. Next, these two maps are concatenated into $F_{\mathrm{cat}}$ so that they can share a convolution layer to keep the module as lightweight as possible.

$$F_{\mathrm{cat}} = [AvgPool_h(F); AvgPool_w(F)] \tag{14}$$

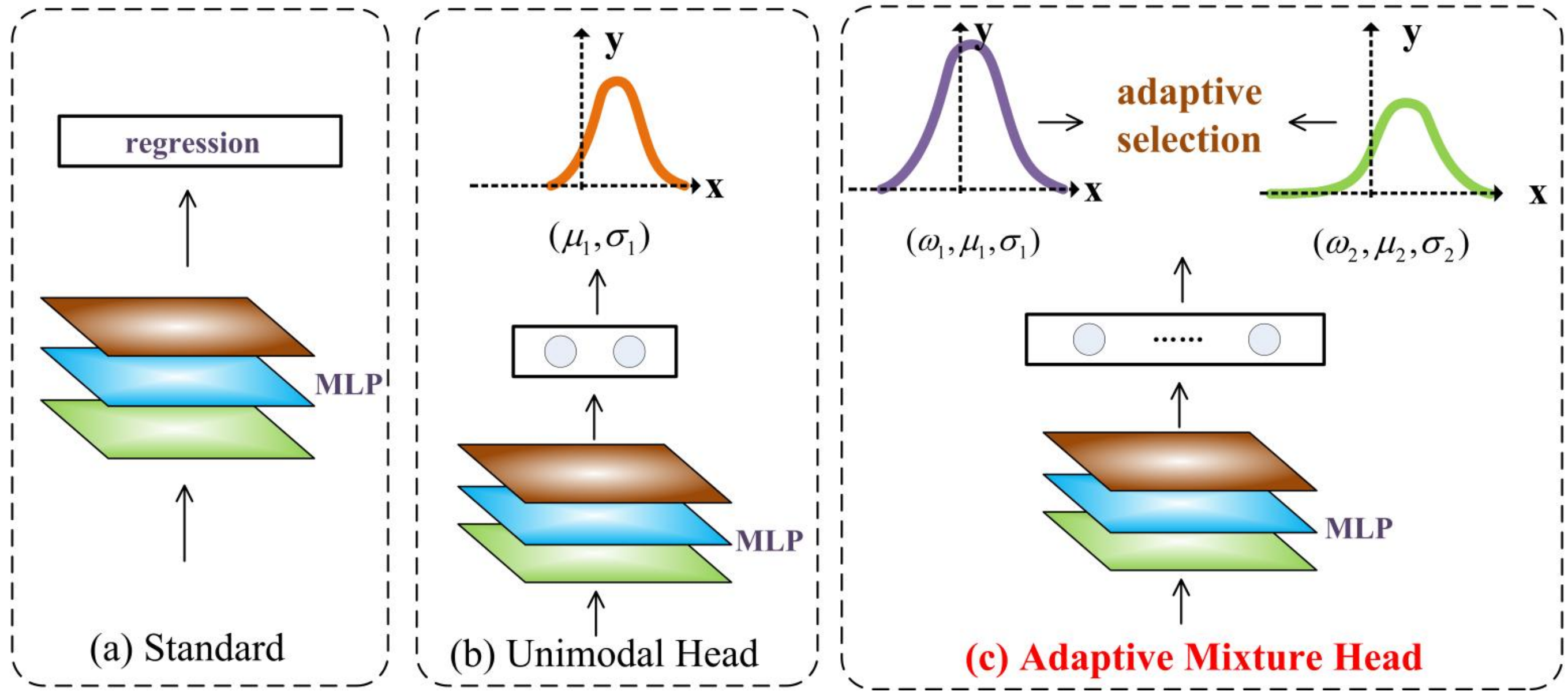

Fig. 7. Comparison of different prediction heads. (a): Standard. (b): Unimodal Head. (c): The proposed adaptive mixture head.

The depthwise separable convolution (a depthwise convolution $Dconv(\cdot)$ followed by a pointwise convolution $Pconv(\cdot)$) [44] is adopted to encode spatial position information of $F_{\text{cat}}$.

$$F_h, F_w = Split[f^{1\times1}(PConv(Dconv(F_{\text{cat}})))] \tag{15}$$

For achieving different cross-channel interactive features between the two parallel paths, after dividing the outputs of 1×1 convolution into two vectors ($Split(\cdot)$), we aggregate the two channel-wise attention maps inside each group via a simple multiplication.

$$w_{p1} = \sigma(f^{1\times1}(F_h)) * \sigma(f^{1\times1}(F_w)) \tag{16}$$

where $\sigma(\cdot)$ is the sigmoid function. In addition, to aggregate multi-scale spatial structure information, a branch starts with a 3 ×3 convolution is placed parallel to the above one. We perform global average pooling ( $GAP(\cdot)$ ) and a $1\times1$ convolution for capturing more distinctive target information from a global perspective. Then, the obtained weights and the output of another $3\times3$ convolution path are multiplied as an attention weight matrix $w_{p2}$ :

$$w_{p2} = \sigma(f^{1\times1}(GAP(F_l))) * \sigma(f^{3\times3}(F_l)) \tag{17}$$

Finally, we obtain the final output feature map through the channel-wise addition and a 1x1 convolution.

$$w = f^{1\times1}(w_{p1} + w_{p2}) \tag{18}$$

*F. Adaptive Mixture Density Head*

For obtaining sharp and precise disparity estimates near discontinuities, SMD-Nets [45] models the disparity using Laplacian distribution and exploits bimodal mixture densities as output representation. However, compared with Laplacian distribution, Gaussian distribution is widely used in various fields due to its convenient mathematical properties [46, 47]. Therefore, in this paper, we use bimodal Gaussian distribution with a novel disparity adaptive selection mechanism, to learn a representation that is precise at object boundaries (Fig. 7(c)).

To be specific, following SMD-Net [45], our backbone network $\psi$ outputs a D-dimensional feature representation with a deterministic transformation:

$$\psi : \mathbb{R}^{W\times H\times 2} \rightarrow \mathbb{R}^D \tag{19}$$

where $D$=96. Note that the features are interpolated bilinear from its nearest four-pixel position for every continuous 2D location.

Next, we adopt a multi-layer perceptron $\Omega$ for mapping the features to a five-dimensional vector $(\omega_1, \mu_1, \mu_2, \sigma_1, \sigma_2)$ , which denotes the parameters of a bimodal Gaussian mixture distribution:

$$\Omega_\theta : \mathbb{R}^D \rightarrow \mathbb{R}^5 \tag{20}$$

The key issue then is to formulate the probability density, and this mixture distribution can be described as:

$$r(d) = \sum_{m=1}^{M} \omega_m \varphi_m(d) \tag{21}$$

where $M$=2 denotes a hyper-parameter denoting the number of components constituting the mixture model. $\omega_m$ represents the mixing coefficient which indicates the probability of different components. $\varphi_m$ denotes the probability density function of component $m$. In this paper, we implement the Gaussian kernel into our framework:

$$\varphi_m(d) = \frac{1}{\sigma_m} e^{-\frac{(\mu_m - d)^2}{2\sigma_m^2}} \tag{22}$$

$$\omega_2 = 1 - \omega_1 \tag{23}$$

where $\sigma_m$ indicates the common variance parameter, and $\mu_m$ denotes the mean of component $m$. By introducing two modes $(\mu_1, \sigma_1)$, $(\mu_2, \sigma_2)$ in our method, we can obtain both the foreground and the background disparity near discontinuities (boundaries).

Therefore, our model can be expressed as:

$$r(d \mid x, I, \theta) = r(d \mid \Omega_\theta(\psi(x, \Psi_\theta(I)))) \tag{24}$$

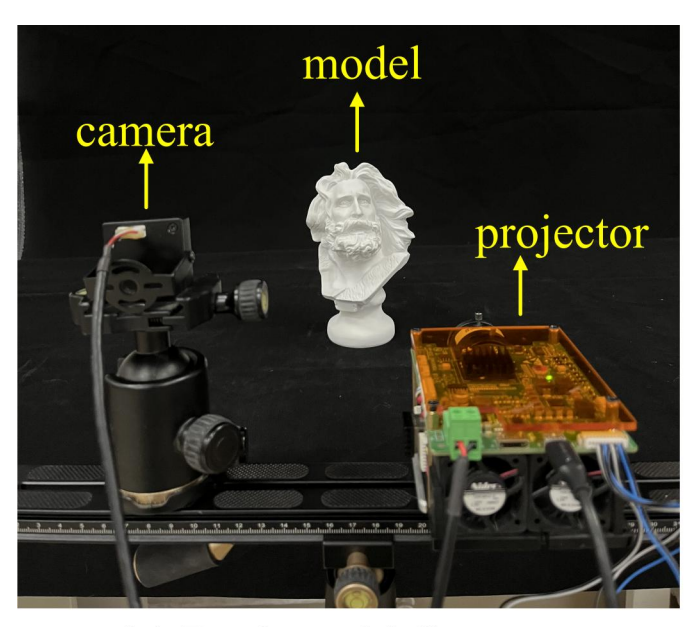

(a) Real-world System

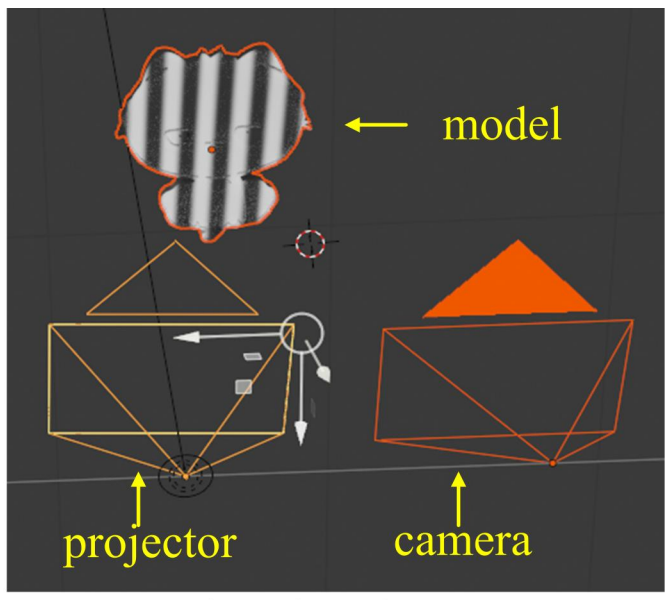

(b) Virtual System

Fig. 8. The experimental systems used in this paper.

TABLE I

DATASETS COMPARISON. N: NON-OCCLUDED. O: OCCLUDED. V: VIRTUAL. R: REAL-WORLD.

| Dataset | Pattern | Scene(Number) | Type |
|---|---|---|---|
| Wang et al. [26] | Fringe | R(600) | N |
| Feng et al. [48] | Fringe | R(960) | N |
| Wang et al. [49] | Fringe | R(1,536) | N |
| Yin et al. [50] | Speckle | R(1,200) | N |
| Ours | Fringe &Speckle | R(1,000) &V(200) | N&O |

where $\Psi_\theta$ is the backbone network, that is, the structure in our network except for the adaptive mixture density head. We refer to $\Omega_\theta(\psi(\cdot,\cdot))$ as our adaptive mixture density head. $\theta$ represents all parameters of the proposed network. $x$ represents a continuous 2D location and $I$ denotes the input image.

At the inference stage, a common strategy is to use the mode with the highest density value to estimate object boundaries. However, this is not robust in practice when the density values of the two modes are close [51]. Therefore, to overcome this problem, we propose a novel adaptive selection strategy. To be specific, we first obtain two candidate disparities by the mode with the highest density value and the lowest density value:

$$\hat{d}_{\max} = \underset{d\in\{\mu_1,\mu_2\}}{\arg\max}\, r(d) \tag{25}$$

$$\hat{d}_{\min} = \underset{d\in\{\mu_1,\mu_2\}}{\arg\min}\, r(d) \tag{26}$$

Then, if the density values corresponding to these two candidate disparities satisfy:

$$\frac{r(d_{\max})}{r(d_{\min})} < thr \tag{27}$$

we choose the average disparity value as the result. If Eq. (27) is not satisfied, we use $d_{\max}$ as the final disparity. In this paper, $thr$=1.1.

### G. Loss Function

We train our base network by minimizing the negative log-likelihood loss:

$$L_{\text{ne}}(\theta) = -\mathbb{E}_{d,x,I} \log r(d \mid x, I, \theta) \tag{28}$$

where $d$ is the ground truth phase at location $x$.

Moreover, for the auxiliary saliency-aware branch, binary cross-entropy loss is used as the loss function to measure the relation between the ground truth and the generated map:

$$L_{\text{sa}} = -\frac{1}{H\times W}\sum_{i=1}^{H}\sum_{j=1}^{W}[T_{ij}\log(P_{ij}) + (1-T_{ij})\log(1-P_{ij})] \tag{29}$$

where $H$ and $W$ are the image's height and width, respectively. $P_{ij}$ represents the probability associated with the presence of salient objects at position $(i, j)$. $T_{ij}$ denotes the ground truth label assigned to the pixel $(i, j)$.

Overall, the total loss is given in Eq (30):

$$L_{\text{to}} = z_{\text{ne}} L_{\text{ne}} + z_{\text{sa}} L_{\text{sa}} \tag{30}$$

where $z$ denotes the weight of different loss functions. In our experiments, $z_{\text{ne}}$=0.8 , $z_{\text{sa}}$=0.2 .

## IV. EXPERIMENTS AND ANALYSIS

### A. Dataset and Implementation Details

*1) Dataset:* We thoroughly review and collect existing SL datasets (Table I), but find that they are either limited to a single modality (e.g., only speckle patterns), contain only simple scenarios (non-occluded), or are not publicly available. To address these limitations, we built a conventional projection profilometry system (Fig. 8(a)) consisting of an industrial camera ($640\times480$) and a DLP projector (LightCrafter 4500). When using the infrared SL system, we divide all the models into 50 groups (including different plaster models, boxes, toys, etc.), where each group of models (including single and multiple models) is captured 20 image pairs (fringes and speckles), and a total of 1000 pairs of images are collected, including different viewing angles and occlusion conditions. We select 40 groups for training and the remaining 10 groups for testing. Note that the models contained in these images are not used during training.

Moreover, to verify the generalization performance, we also use the Blender software to build a virtual SL system (Figure 8(b)) and generate 200 image pairs using different models in a similar manner. Specifically, 10 image pairs are captured for each group of models, resulting in a total of 200 pairs. All of these images are used for direct testing. Note that the 3D models used for this system are obtained from free websites. Complementary Gray-code-based phase unwrapping method [52] is exploited to obtain the ground-truth phase maps because it demonstrates superior anti-interference ability compared to other unwrapping methods [53]. The speckle pattern is captured from Intel RealSense.

*2) Metrics:* We use two metrics: the End Point Error (EPE), which is defined as the minimum absolute error between the estimated and ground-truth phase; the $x$-pixels error (ERR$x$), which is the percentage of pixels having errors greater than $x$.

*3) Implementation Details:* Our experiments are carried out on a machine with two NVIDIA GeForce RTX 4090 GPUs. The number of epochs for training is set to 500, and the batch size is set to 2. We compare PDCNet with four single-shot state-of-the-art methods: FOA-Net [9], CTransU-Net [10], LightUNet [23], and PCTNet [11]. Among these networks,

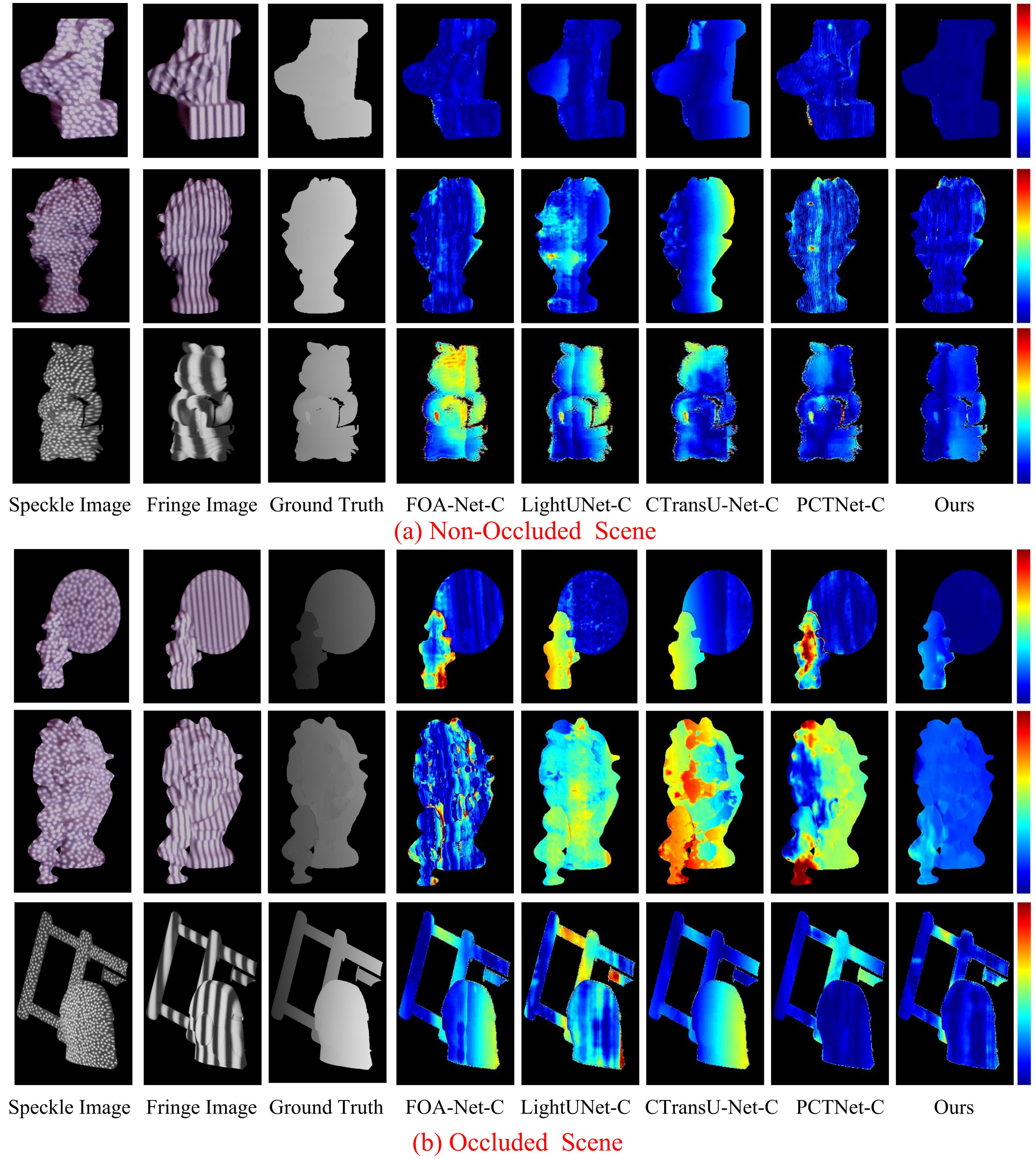

Fig. 9. Qualitative comparison results. The fourth column to the eighth column show the error maps obtained from different methods. The third row shows the test results on the virtual dataset. The suffix C denotes the input is a concatenation of fringe images and the corresponding speckle images.

TABLE II

QUANTITATIVE RESULTS ON REAL-WORLD DATASET. THE SUFFIX C DENOTES THE INPUT IS A CONCATENATION OF FRINGE IMAGES AND THE CORRESPONDING SPECKLE IMAGES.

| Method | Framework | EPE | ERR3 | ERR5 | ERR10 |
|---|---|---|---|---|---|
| FOA-Net | *C* | 1.67 | 14.43 | 10.40 | 4.43 |
| LightUNet | C | 1.84 | 16.86 | 5.00 | 1.93 |
| CTransU-Net | *T* | 1.97 | 18.92 | 5.65 | 1.91 |
| PCTNet | *C* + *T* | 1.92 | 16.62 | 5.97 | 1.83 |
| FOA-Net-C | *C* | 1.36 | 9.47 | 6.01 | 3.53 |
| LightUNet-C | C | 1.39 | 10.28 | 6.33 | 1.67 |
| CTransU-Net-C | *T* | 1.42 | 13.45 | 4.93 | 1.94 |
| PCTNet-C | *C* + *T* | 1.37 | 9.71 | 5.27 | 1.70 |
| PDCNet (Ours) | *C* // *T* | **0.98** | **6.91** | **3.37** | **1.20** |

TABLE III

QUANTITATIVE RESULTS ON VIRTUAL DATASET.

| Method | Framework | EPE | ERR3 | ERR5 | ERR10 |
|---|---|---|---|---|---|
| FOA-Net-C | *C* | 7.40 | 19.13 | 15.71 | 7.25 |
| LightUNet-C | *C* | 9.87 | 22.46 | 16.90 | 8.60 |
| CTransU-Net-C | *T* | 9.55 | 23.80 | 18.04 | 8.76 |
| PCTNet-C | *C*+*T* | 5.62 | 15.77 | 10.52 | 6.42 |
| PDCNet (Ours) | *C* | **4.03** | **15.52** | **9.37** | **5.17** |

TABLE IV

COMPARISON OF DIFFERENT ARCHITECTURES ON REAL-WORLD DATASET.

| Architecture | Fringe / Speckle | EPE | ERR3 | ERR5 | ERR10 |
|---|---|---|---|---|---|
| a) | CNN / CNN | 1.23 | 8.63 | 4.90 | **1.18** |
| b) | CNN / Trans | 1.34 | 10.69 | 7.02 | 3.96 |
| c) | Trans / Trans | 1.27 | 8.81 | 4.57 | 2.13 |
| d) | Trans / CNN (Ours) | **0.98** | **6.91** | **3.37** | 1.20 |

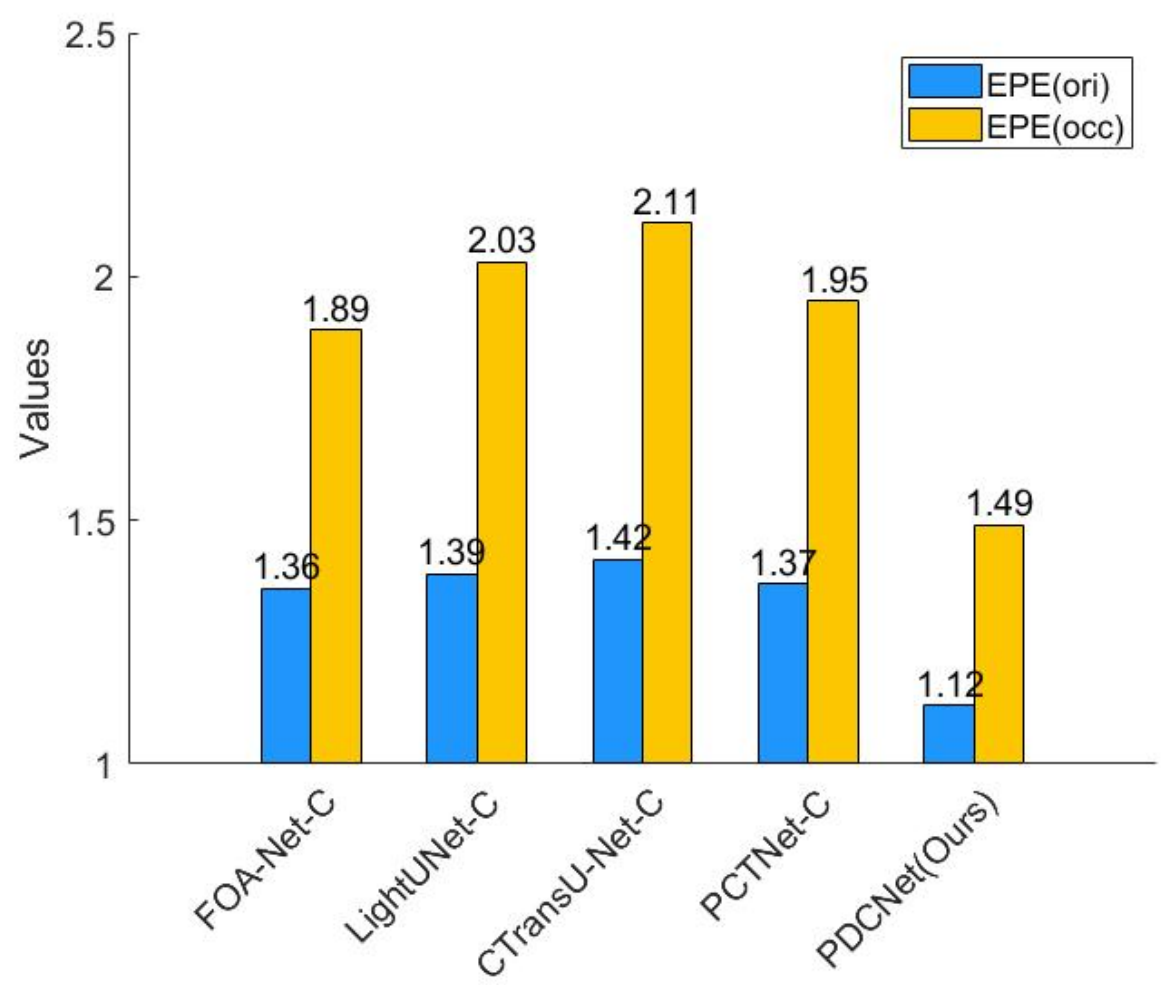

Fig. 10. Comparison of different methods on the original real-world test set and the occluded-only real-world test set.

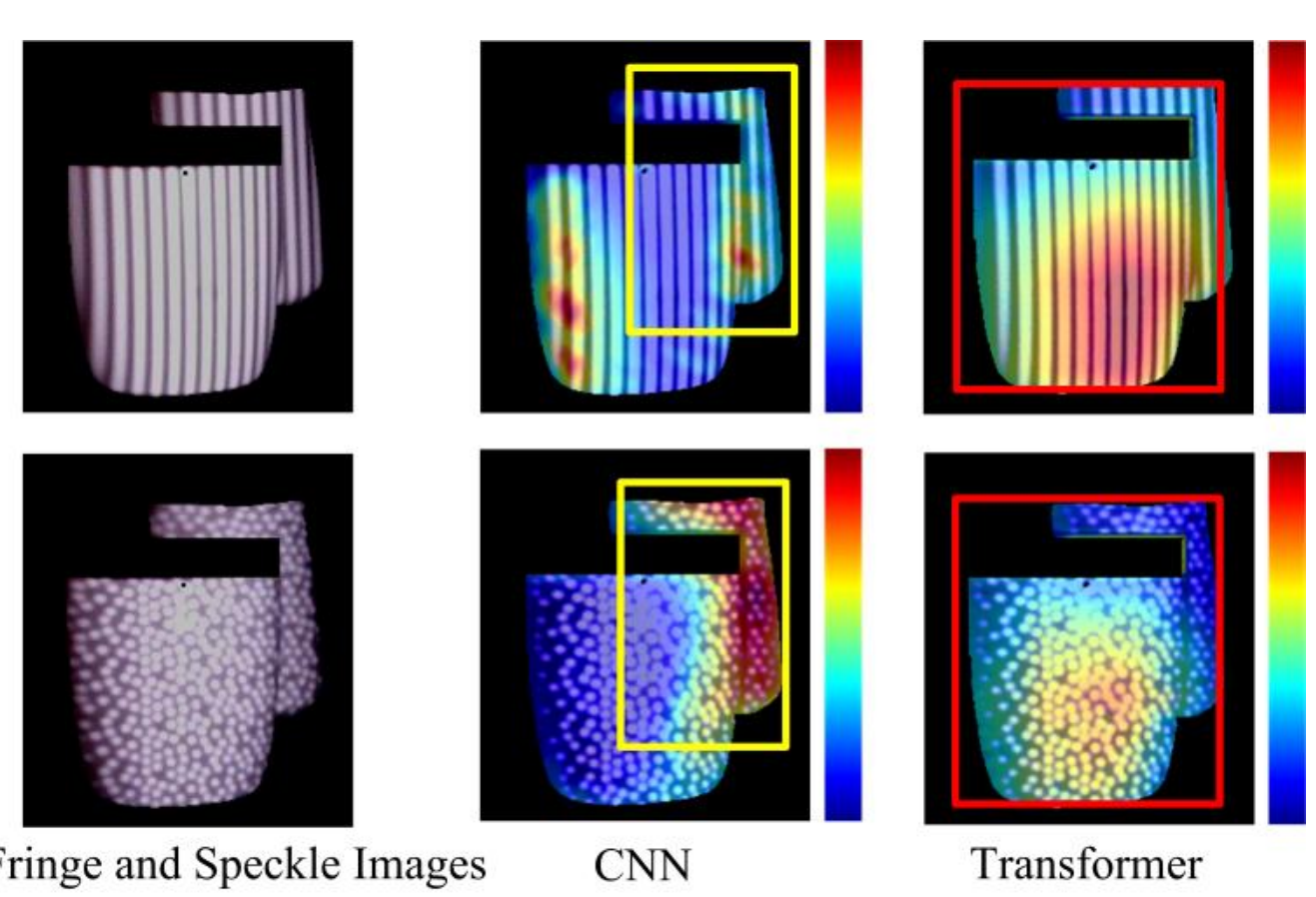

Fig. 11. Comparison of heatmaps of different branches. The first column shows the input images.

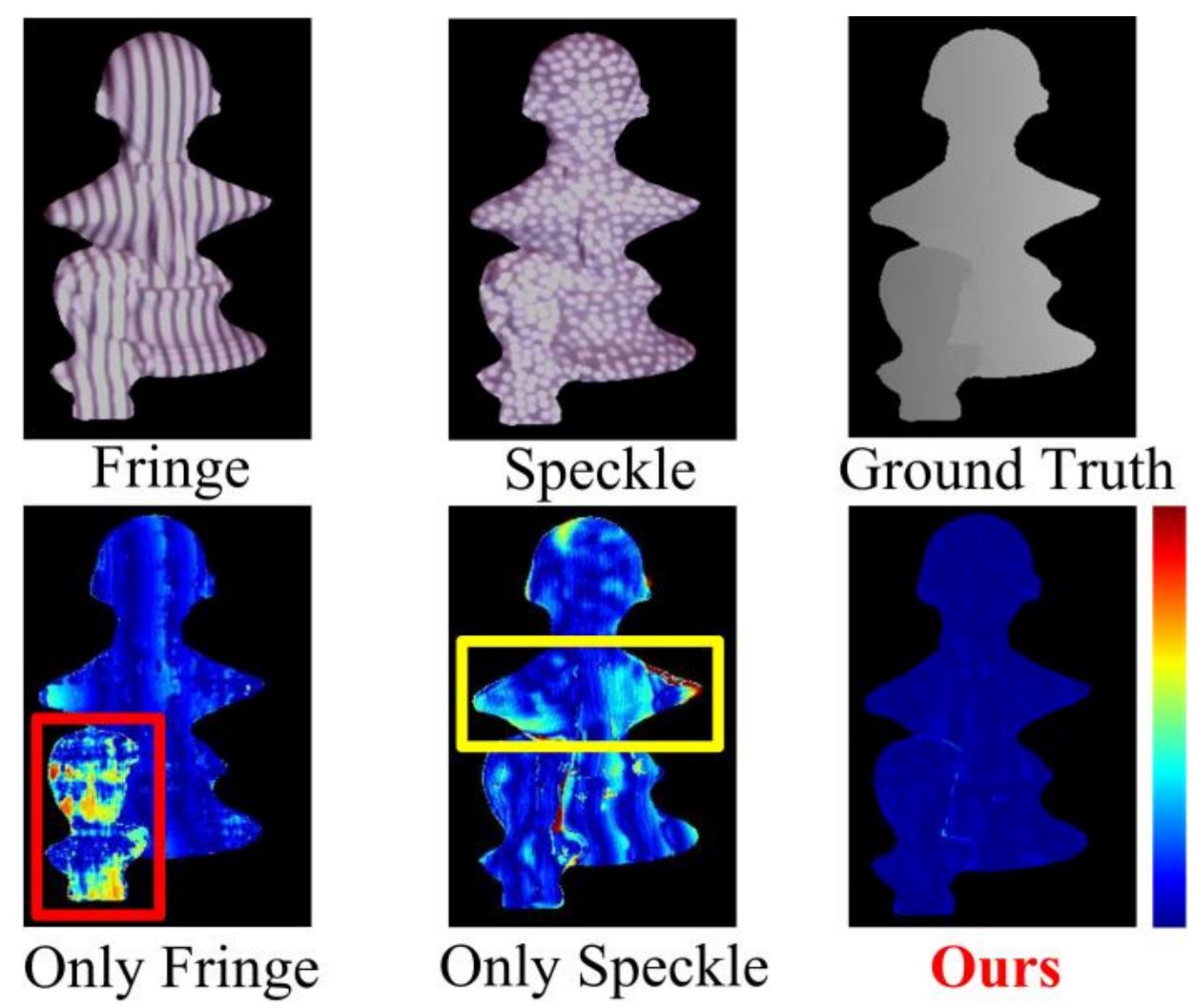

Fig. 12. Comparison with single-modal architectures. The second row shows the error maps of different architectures.

TABLE V

COMPARISON OF DIFFERENT $z_{ne}$ ON REAL-WORLD DATASET.

| $z_{ne}$ | EPE | ERR3 |
|---|---|---|
| 0.5 | 1.21 | 7.83 |
| 0.6 | 1.20 | 7.96 |
| 0.7 | 1.13 | 7.12 |
| 0.8 | **0.98** | **6.91** |
| 0.9 | 1.15 | 7.06 |

TABLE VI

COMPARISON OF DIFFERENT DEPTH CONFIGURATIONS ON REAL-WORLD DATASET.

| $N_{LT}$ | $N_{HP}$ | EPE |
|---|---|---|
| 1 | 1 | 1.20 |
| 2 | 2 | 1.14 |
| 3 | 3 | **0.98** |
| 4 | 4 | 1.14 |
| 5 | 5 | 1.15 |

the first two use a fringe image for measurement. Moreover, we also concatenate the fringe images and the corresponding speckle images as input for retraining and retesting. In Table II, they are denoted by the suffix C. Note that in this paper, we fine-tune the input and output layers of these networks to fit our task.

### *B. Qualitative Evaluation*

The phase errors of different methods are shown in Fig. 9. We can observe that our method better integrates global structures in fringe images and local details in speckle images. To be specific, when only isolated objects are present in the non-occluded scenes, the performance of various methods shows little difference. In contrast, when objects overlap with each other in the occluded scenes, our method demonstrates better perceptual capabilities and shows strong adaptation power.

Moreover, Fig. 9 demonstrates that the measurement results of our PDCNet are close to those obtained by traditional method (ground-truths). However, our method requires fewer input images and eliminates the need for manual fine-tuning, thus having the potential to enhance measurement efficiency in practical applications.

### *C. Quantitative Evaluation*

In Table II and Table III, four metrics are used to quantitatively compare with other methods. Our method has excellent performance on all metrics, demonstrating that our PDCNet is suitable for various target categories, which indicates the effectiveness of our parallel dual-branch CNN-Transformer framework. We can also observe that the average error on the virtual test set is higher than that on the real-world test set. This can be attributed to that these existing deep learning-based methods generally learn all the features in the SL images without distinguishing between SL and non-SL information. If the model processes all the features together, it can be difficult to perceive subtle changes in SL patterns across different scenes. This can lead to inaccurate depth estimation and affect the measurement performance. Therefore, our future research direction is SL image feature decomposition. Fig. 10 shows the performance of the different methods on the original test set and the occluded-only test set, and it can be seen that our method has a more obvious advantage on the latter.

### *D. Ablation Studies*

*1) Effect of Different Architectures.* We verify the necessity of LTB and the HPB in the corresponding branch. The results are shown in Table IV. In architecture a), we change LTB as HPB, that is, both the fringe and speckle features are extracted by HPB. Similarly, in architecture c), these features are both extracted by LTB. The results demonstrate that although the ability of feature extraction for HPB is stronger than that of LTB, it is still worse than that of our PDCNet which cooperates with HPB and LTB. In architecture b), we use HPB and LTB to extract fringe and speckle features respectively, and its effect is also worse than that of our method, which proves that the proposed architecture effectively leverages the advantages of both modalities.

We also visualize the output heatmaps of different architectures in Fig. 11. We can observe that: 1) When using CNN to process speckle images, the local regions prone to fringe ambiguity (yellow boxes) are paid more attention. 2) When using Transformer to process fringe images, the global information (red boxes) is captured more comprehensively.

We believe the reason for the first point lies in the local receptive fields in CNN. Speckle images contain stable speckle distribution patterns, and CNN can gradually extract higher-level features, allowing for better understanding and differentiation of important local structures (yellow boxes) in speckle images.

TABLE VII

EVALUATION OF DIFFERENT ATTENTION MECHANISMS ON THE ORIGINAL REAL-WORLD TEST SET AND THE OCCLUDED-ONLY REAL-WORLD TEST SET.

| Mechanism | EPE (original) | EPE (occluded) |
| --- | --- | --- |
| DAAM (Ours) | **0.98** | **1.37** |
| CBAM | 1.06 | 1.59 |
| SE | 1.10 | 1.65 |

TABLE VIII

EVALUATION OF SALIENCY-AWARE PATHS ON REAL-WORLD DATASET.

| Path | EPE | ERR3 |
| --- | --- | --- |
| Ours | **0.98** | **6.91** |
| Ours-BT | 1.18 | 7.64 |
| Ours-BC | 1.21 | 7.70 |
| Ours-BT-BC | 1.25 | 8.32 |

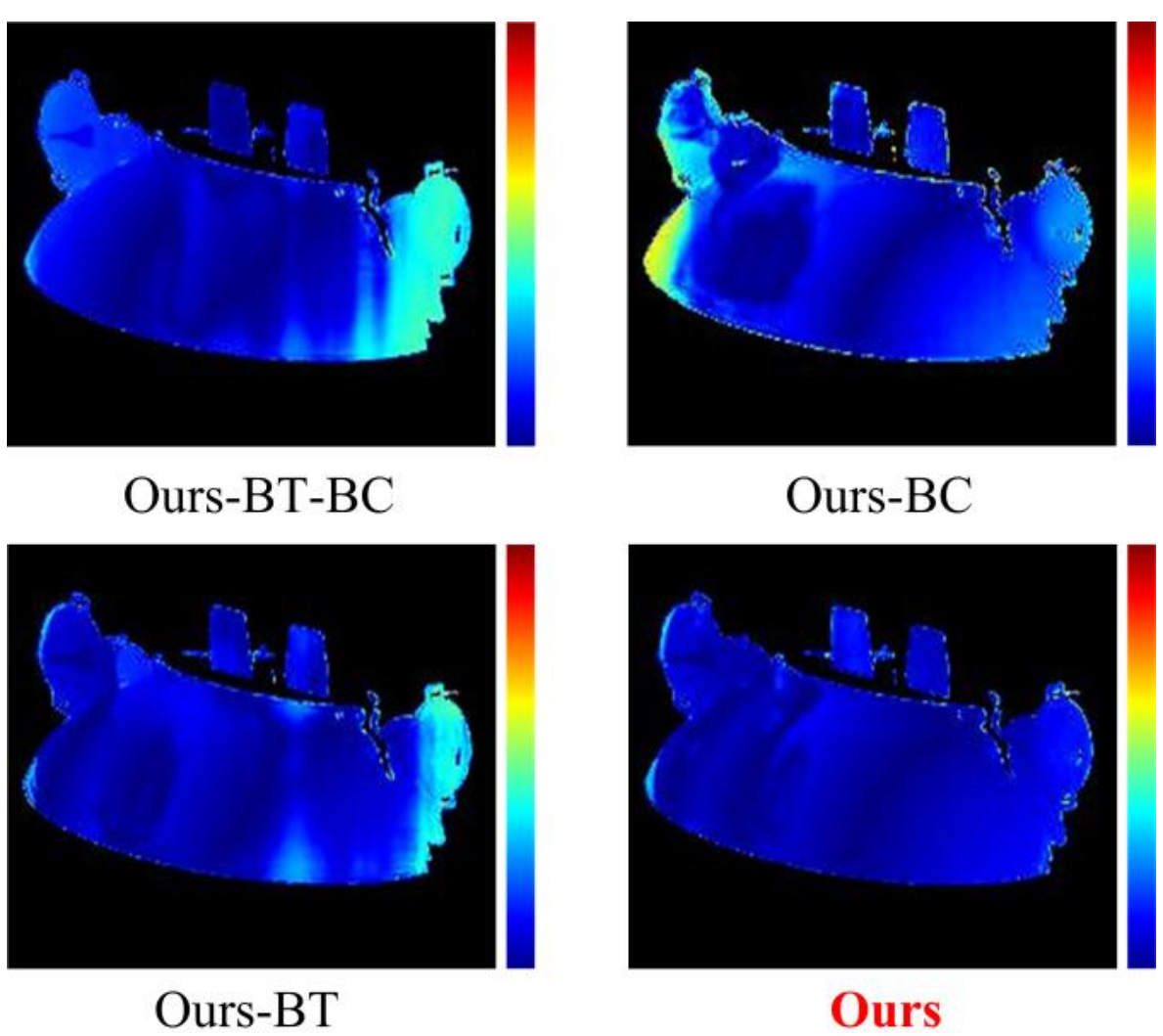

Fig. 13. Comparison of error maps of different inputs to the saliency-aware path.

On the other hand, the reason for the second point lies in the self-attention mechanism in Transformer. Fringe images typically exhibit long-range correlations, such as fringes extending across the entire image region (red boxes). The self-attention mechanism models enable them to establish long-distance dependencies between different positions, thus better capturing global features and contextual information in images. Moreover, we also capture single-modal features using HPB (only speckle) and LTB (only fringe), respectively. As shown in Fig. 12, the results indicate that the single modality achieves worse performance compared with the double modalities. While the fringe image encodes sinusoidal patterns pixel by pixel, significant depth gaps between adjacent objects can lead to ambiguous fringe orders, causing inaccurate predictions (red box). On the other hand, although speckle images exhibit clear spatial features unaffected by depth changes, their lack of pixel-by-pixel encoding may generate poor results in practical measurements (yellow box). Our network can promote these complementary features.

*2) Effect of Hyperparameters.* Within our PDCNet, $z_{ne}$ in the combined loss function is a hyperparameter to balance the base loss and the auxiliary saliency-aware loss. As shown in Table V, extensive experiments show that the best performance is when $z_{ne}$=0.8. Moreover, we compare the effects of different depth configurations in Table VI. Let $N_{LT}$ and $N_{HP}$ indicate the repeating number of the LTB and the HPB in the corresponding branch respectively. We notice that simply increasing the depth would not always bring better performance. When the repeating number in each branch is set to 3, the localization error is minimal.

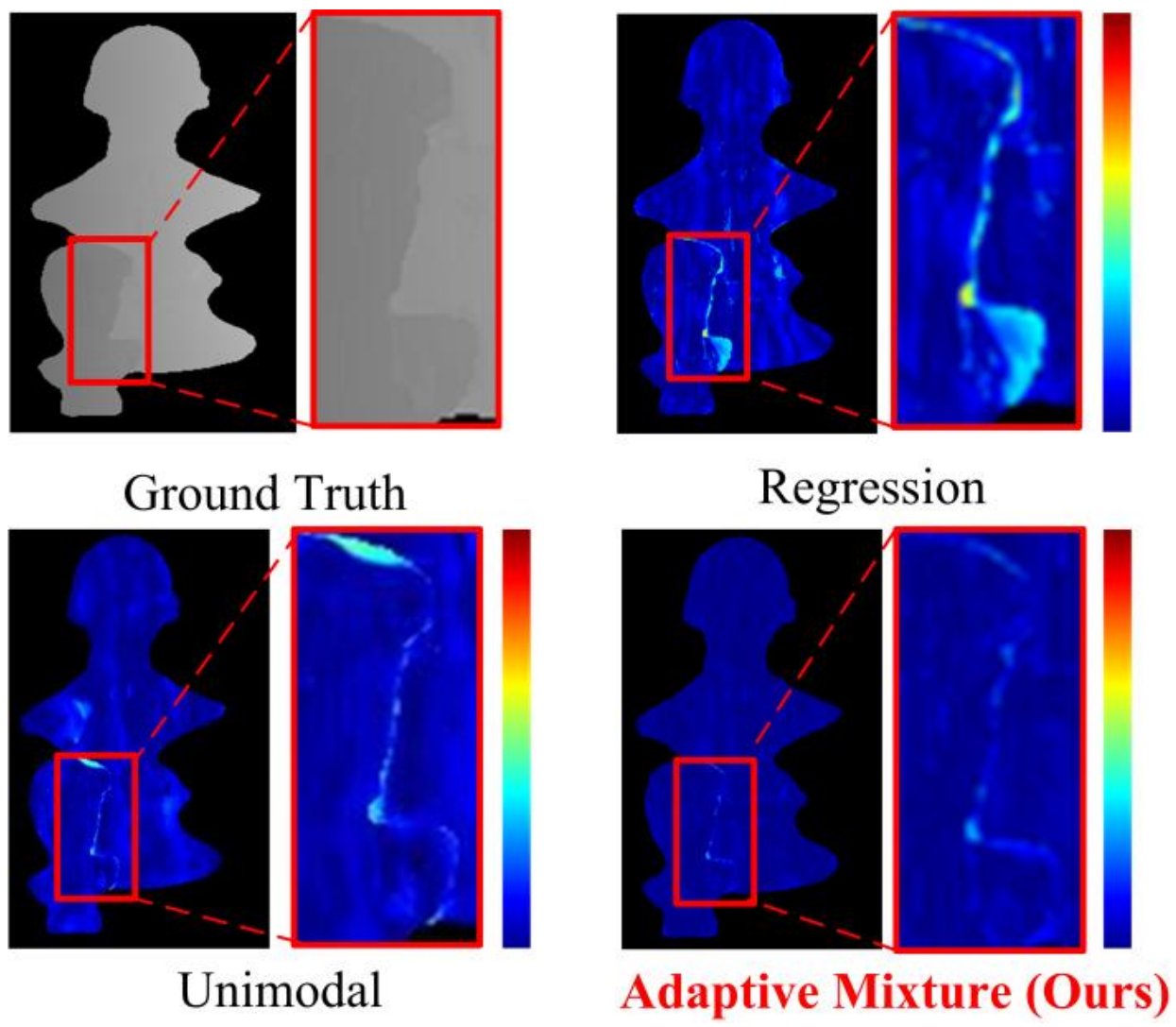

Fig. 14. Comparison of error maps of different prediction heads.

TABLE IX

EVALUATION OF DIFFERENT PREDICTION HEADS ON REAL-WORLD DATASET.

| Head | EPE | ERR3 | SEE |
| --- | --- | --- | --- |
| Regression | 1.12 | 7.00 | 1.77 |
| Unimodal | 1.03 | **6.89** | 1.46 |
| Adaptive Mixture | **0.98** | 6.91 | **1.25** |

*3) Effect of DAAM.* We compare our DAAM with other lightweight attention modules [54-56], including the SE attention and CBAM. For other settings, we follow the original PDCNet. As shown in Table VII, our DAAM consistently demonstrates superior performance across both the original and occluded-only test sets and exhibits more significantly superior results on the latter. Specifically, DAAM surpasses SE attention by an average of 13.93% on precision. We argue that the advantage benefits from essential location information, whereas SE attention only takes information between channels into account. Moreover, DAAM outperforms CBAM by an average of 10.67% on precision. We think this is because CBAM utilizes a convolutional layer to encode local spatial information while DAAM encodes global information with two complementary 1D global pooling operations, which enables our attention to capture long-range dependencies. These factors are the reason why our modules also perform well on the occluded-only test sets.

*4) Effect of Saliency-Aware Path.* Table VIII shows the results after removing the saliency-aware path in the Transformer branch (denoted as BT) and CNN branch (denoted as BC). Despite this branch is used only during the training stage, it can lead to at least a 5.08% performance improvement. From Figure 13, we can also observe that when the information of both branches is input into the saliency-aware path, the measurement error is lowest. This scheme, which does not introduce additional costs during the inference stage, can also be applied to other tasks involving SL images.

*5) Effect of Adaptive Mixture Density Head.* The choice of output representation will directly affect the final performance. Therefore, we explore the impact of different representations in Table IX. In addition to EPE, we also evaluate the Soft Edge Error (SEE) metric on pixels belonging to boundaries [45]. We can observe that the best depth estimation performance is achieved when the proposed adaptive mixture density head is used, outperforming other representations. It is noteworthy that on the self-made dataset, the unimodal representation performs worse than the standard regression, exhibiting higher EPE and SEE. This also highlights the instability of the unimodal representation in the depth estimation task for speckle images. Fig. 14 presents that the proposed mixture head effectively improves performance at object boundaries, compared to both the standard disparity regression and the unimodal representation.

## V. Conclusion

In this paper, we propose an effective end-to-end dual-branch CNN-Transformer framework for 3D shape measurement. Specifically, we leverage the CNN branch to extract local information from speckle images and the Transformer to obtain global representations from fringe images. In addition, we design a double-stream attention aggregation module to integrate complementary features from different branches. An adaptive mixture density head with bimodal Gaussian distribution is used for learning a representation that is precise near discontinuities. Extensively comparative experiments reveal the advantage of our PDCNet over state-of-the-art deep learning-based measurement algorithms in terms of qualitative and quantitative assessment. In the future, we will try to test the proposed framework on different tasks, as it can process images with global and local information respectively and integrate complementary features. In addition to speckle-fringe image pairs, visible-infrared image pairs and MRI-PET image pairs also have similar characteristics.

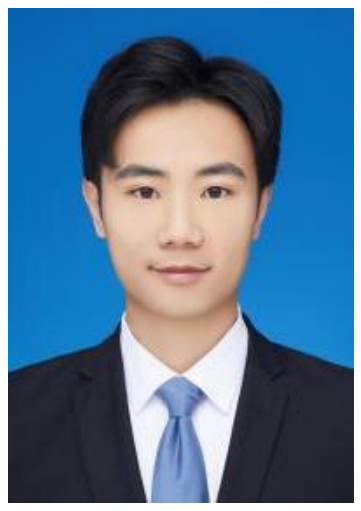

**Mingyang Lei** received the M.S. degree from North China University of Technology, China, in 2020. He is now pursuing the Ph.D. degree with the School of Medical Technology, Beijing Institute of Technology, China. His research interests include remote sensing image analysis, deep learning and computer vision.

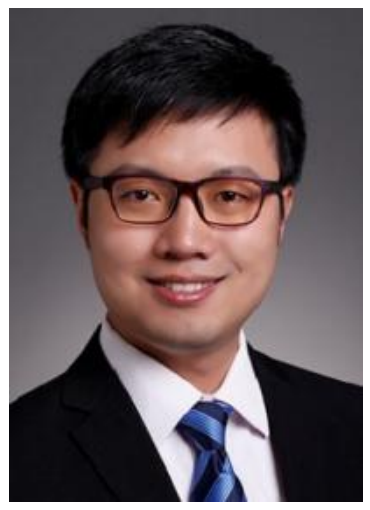

**Jingfan Fan** received the Ph.D. degree in optical engineering from Beijing Institute of Technology, China in 2016, and worked as post-doctoral researcher with University of North Carolina at Chapel Hill in US from 2017 to 2019. He is currently an associate professor with the School of Optics and Photonics, Beijing Institute of Technology. He focuses his research interests on medical image processing, computer vision and augmented reality.

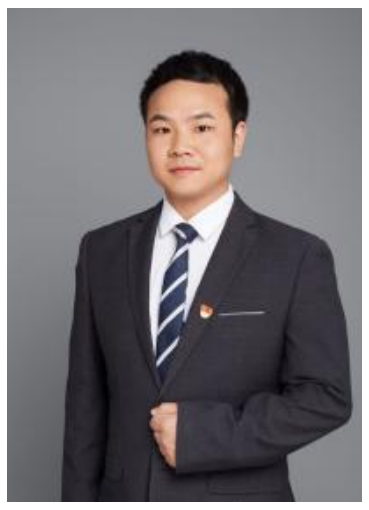

**Long Shao** received the B.E. degree in measurement and control technology and instrumentation from Xi 'an Technological University in 2015, and the ph.D. degree in optical engineering from Beijing Institute of Technology in 2022. He is a postdoctoral fellow in the School of Computer Science and Technology, Beijing Institute of Technology, Beijing, China. His research interests include binocular stereo vision, intelligent image sensing, and computer vision.

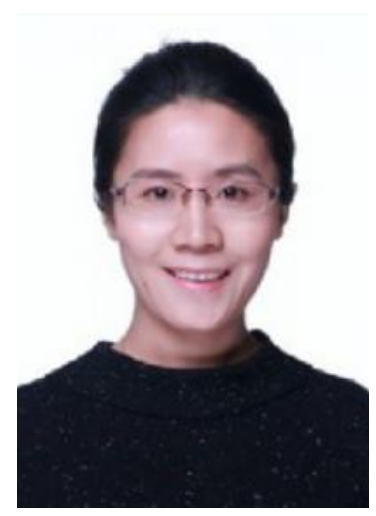

**Hong Song** received the Ph.D. degree in computer science from Beijing Institute of Technology, China in 2004. She is currently a Professor with the School of Computer Science and Technology, Beijing Institute of Technology. Her research interests focus on medical image processing, augmented reality and computer vision..

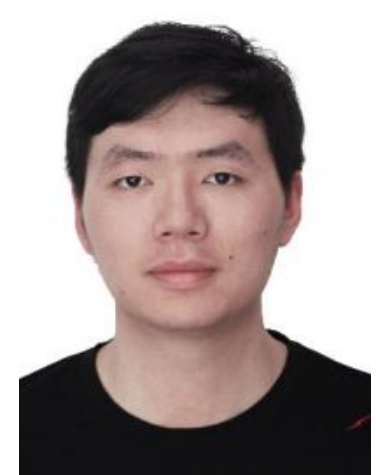

**Deqiang Xiao** received the B.S. degree in computer science from Henan University, Kaifeng, China, in 2011, and the Ph.D. degree from the Shenzhen Institutes of Advanced Technology, University of Chinese Academy of Sciences, Shenzhen, China, in 2017. He is currently an Assistant Professor with the School of Optics and Photonics, Beijing Institute of Technology, Beijing, China. His research interests include medical image analysis, image-guided intervention/surgery, and computer vision.

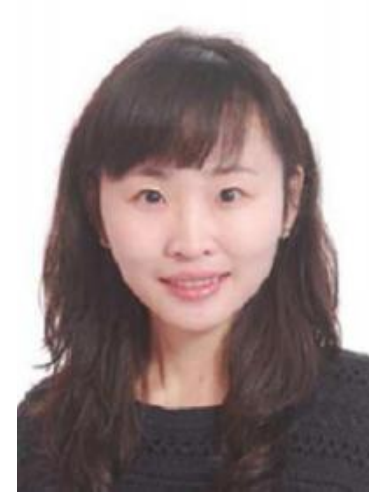

**Danni Ai** received the Ph.D. degree from Ritsumeikan University, Japan in 2011. She is currently an associate professor with the School of Optics and Photonics, Beijing Institute of Technology, China. Her research interests include medical image analysis, surgical navigation, virtual reality and augmented reality.

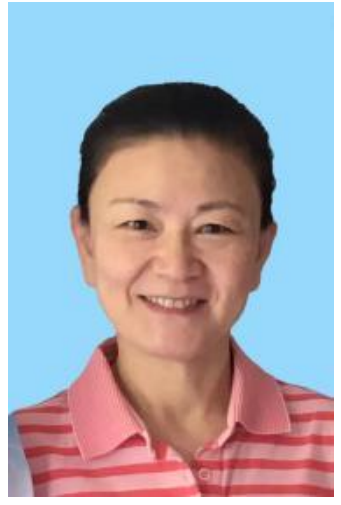

**Ying Gu** received the Ph.D. degree from Chinese People's Liberation Army (PLA) Medical School China in 2000. She is currently an chief physician and professor of Department of Laser Medicine, the First Center of Chinese PLA General Hospital & Medical School of Chinese PLA, and academician of Chinese Academy of Sciences, China. Her research interests include vascular targeted photodynamic therapy (V-PDT), medical image analysis, and surgical navigation.

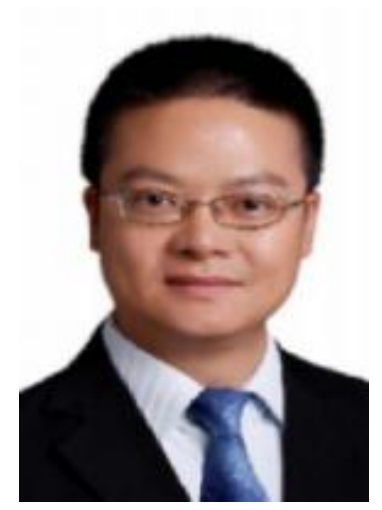

**Jian Yang** received the Ph.D. degree in optical engineering from Beijing Institute of Technology, China in 2007. He is currently a professor with the School of Optics and Photonics, Beijing Institute of Technology. His research interests focus on medical image processing, augmented reality and computer vision.